%% file: main.tex
\documentclass[letterpaper, 10 pt, journal, twoside,table,xcdraw]{IEEEtran}  
\IEEEoverridecommandlockouts                              
                           
\usepackage{balance} 
\usepackage[pdftex]{graphicx}
\usepackage[footnotesize]{caption}
\input{header.tex}

\usepackage{tikz}
\usepackage{color}

\definecolor{darkpastelgreen}{rgb}{0.01, 0.75, 0.24}
\usetikzlibrary{matrix,calc}
\tikzstyle{block} = [draw,rectangle,thick,minimum height=2em,minimum width=2em]
\tikzstyle{arrow} = [->,thick]
\usepackage{bm}

\usepackage{multirow}
\usepackage{xcolor}
\usepackage[normalem]{ulem}
\useunder{\uline}{\ul}{}

\makeatletter
\newcommand\fs@ruled@notop{\def\@fs@cfont{\bfseries}\let\@fs@capt\floatc@ruled
  \def\@fs@pre{}%
  \def\@fs@post{\kern2pt\hrule\relax}%
  \def\@fs@mid{\kern2pt\hrule\kern2pt}%
  \let\@fs@iftopcapt\iftrue}
\renewcommand\fst@algorithm{\fs@ruled@notop}
\makeatother
\allowdisplaybreaks

\begin{document}

\title{Efficient Automatic Tuning for Data-driven \\ Model Predictive Control via Meta-Learning}
\author{
Baoyu Li\textsuperscript{1}\thanks{\noindent \textsuperscript{1}B. Li, W. Edwards, and K. Hauser are with the Department of Computer Science, University of Illinois at Urbana-Champaign, IL, USA. 
\texttt{\{baoyul2,wre2,kkhauser\}@illinois.edu}}, 
\and William Edwards\textsuperscript{1},
\and Kris Hauser\textsuperscript{1}
   }

\setlength\abovedisplayskip{3pt}
\setlength\belowdisplayskip{3pt}

\maketitle

\begin{abstract}

\setcounter{footnote}{1}
\texttt{AutoMPC}\footnote{The code and documentation for \texttt{AutoMPC} and its improved version are available at \url{https://github.com/uiuc-iml/autompc}.} is a Python package that automates and optimizes data-driven model predictive control. However, it can be computationally expensive and unstable when exploring large search spaces using pure Bayesian Optimization (BO). To address these issues, this paper proposes to employ a meta-learning approach called \textit{Portfolio} that improves \texttt{AutoMPC}'s efficiency and stability by warmstarting BO. Portfolio optimizes initial designs for BO using a diverse set of configurations from previous tasks and stabilizes the tuning process by fixing initial configurations instead of selecting them randomly. Experimental results demonstrate that Portfolio outperforms the pure BO in finding desirable solutions for \texttt{AutoMPC} within limited computational resources on 11 nonlinear control simulation benchmarks and 1 physical underwater soft robot dataset.

\end{abstract}




\input{01-introduction}
\input{02-related}
\input{03-methods}

\input{04-experiments}
\input{05-conclusion}



\clearpage
\bibliographystyle{IEEEtran}
\bibliography{references}

\clearpage
\input{06-appendix}
\end{document}

%% file: header.tex
\usepackage{amsmath,amsfonts,amssymb}
\usepackage{textgreek}
\usepackage{tabu}
\usepackage{bbm}
\usepackage{prettyref}
\usepackage{mathrsfs}
\usepackage{graphicx}
\usepackage{wrapfig}
\usepackage{subcaption}
\usepackage{MnSymbol}
\usepackage{multirow}
\usepackage{booktabs}
\usepackage{algorithm}
\usepackage[table]{xcolor}
\usepackage{cellspace}
\usepackage{mathtools}
\usepackage{cite}
\usepackage{pbox}
\usepackage{hyperref}
\usepackage{xfrac}
\usepackage{soul}
\usepackage{comment}
\usepackage{algorithmic}

\newcommand*{\colorboxed}{}
\def\colorboxed#1#{%
  \colorboxedAux{#1}%
}
\newcommand*{\colorboxedAux}[3]{%
  \begingroup
    \colorlet{cb@saved}{.}%
    \color#1{#2}%
    \boxed{%
      \color{cb@saved}%
      #3%
    }%
  \endgroup
}

\newrefformat{Fig}{Figure~\ref{#1}}
\newrefformat{fig}{Figure~\ref{#1}}
\newrefformat{par}{Section~\ref{#1}}
\newrefformat{appen}{Appendix~\ref{#1}}
\newrefformat{ass}{A~\ref{#1}}
\newrefformat{sec}{Section~\ref{#1}}
\newrefformat{sub}{Section~\ref{#1}}
\newrefformat{table}{Table~\ref{#1}}
\newrefformat{alg}{Algorithm~\ref{#1}}
\newrefformat{Alg}{Algorithm~\ref{#1}}
\newrefformat{Def}{Definition~\ref{#1}}
\newrefformat{Cor}{Corollary~\ref{#1}}
\newrefformat{Thm}{Theorem~\ref{#1}}
\newrefformat{Lem}{Lemma~\ref{#1}}
\newrefformat{step}{Step~\ref{#1}}
\newrefformat{ln}{Line~\ref{#1}}
\newrefformat{eq}{Equation~\ref{#1}}
\newrefformat{pb}{Problem~\ref{#1}}
\newrefformat{it}{Item~\ref{#1}}
\newrefformat{te}{Term~\ref{#1}}
\def\Eqref Eq:#1:{\eqref{eq:#1}}
\newrefformat{Eq}{Equation~\Eqref#1:}

\newcommand{\R}{\mathbb{R}}

\newcommand{\E}[1]{\mathbf{#1}}
\newcommand{\TE}[1]{\textbf{#1}}

\newcommand{\argminP}[1]{\E{argmin}\;}

\newcommand{\argmaxP}[1]{\E{argmax}\;}

%% file: 01-introduction.tex
\section{Introduction}

Data-driven model predictive control (MPC) is a robust and flexible framework for robotic controller design~\cite{bemporad1999robust}. By leveraging knowledge of the data-learned dynamics system, it can predict a robot’s behavior over a specified time horizon and determines the optimal control actions that will achieve the desired objectives while satisfying constraints. Due to its capacity to handle intricate nonlinear systems, constraints, and uncertainties in real-time, data-driven MPC has broad applications in robotics, including trajectory tracking~\cite{abdollahyan2020data}, obstacle avoidance~\cite{palma2020data}, and manipulation tasks~\cite{ortiz2021data}. 

However, successfully implementing data-driven MPC requires expert knowledge to make various design choices such as objective functions, system identification (SysID), and optimization techniques. Additionally, simple data-driven MPC necessitates intricate manual hyperparameter tuning for these design components, which can prove to be time-consuming and susceptible to errors.

To address these issues, the research community has made efforts to democratize data-driven MPC for users with no expert-level knowledge in the delicate MPC design process~\cite{bansal2017goal, piga2019performance, edwards2021automatic}. Most recently, Edwards et al.~\cite{edwards2021automatic} proposed an end-to-end automatic tuning pipeline for data-driven MPC, named \texttt{AutoMPC}, which optimizes hyperparameters by Bayesian Optimization (BO) and evaluates performance via a cross-validation-like technique using surrogate dynamics. However, \texttt{AutoMPC} has a significant limitation: it begins from random initialization on every new problem and does not leverage knowledge from previous encountered tasks, leading to unstable tuning processes and sub-optimal performance for dynamics modeling and control.


Inspired by recent advances in AutoML~\cite{feurer-autosklearn, JMLR-autokeras}, we propose to apply an alternate meta-learning-based paradigm called \textit{Portfolio} in this paper. Portfolio allows \texttt{AutoMPC} to adjust to new tasks rapidly, similar to how it warms up the AutoML system. It provides an optimized initial design for pure BO, drawn a diverse and dependable set of configurations from previous runs. 
By replacing the random initialization of pure BO with a method which leverages prior knowledge, Portfolio improves the stability and effectiveness of the tuning process. We employ portfolio-based meta-learning on model tuning, which serves as a major component in \texttt{AutoMPC} pipeline. 

We conduct extensive experiments on various benchmarks to build a comprehensive Portfolio and validate the contributions of our proposed method. Besides the three benchmarks described in~\cite{edwards2021automatic}, we perform experiments on more nonlinear control benchmarks from the OpenAI Gym MuJoCo~\cite{gym}, Gym extensions~\cite{henderson2017multitask}, and underwater soft robot~\cite{NullSoftRobot2022, null2023automatic}. Experimental results on 11 nonlinear control simulation benchmarks and 1 dataset collected from physical underwater soft robot demonstrate the efficacy of Portfolio.

\begin{figure*}
    \centering
    \includegraphics[width=0.8\textwidth]{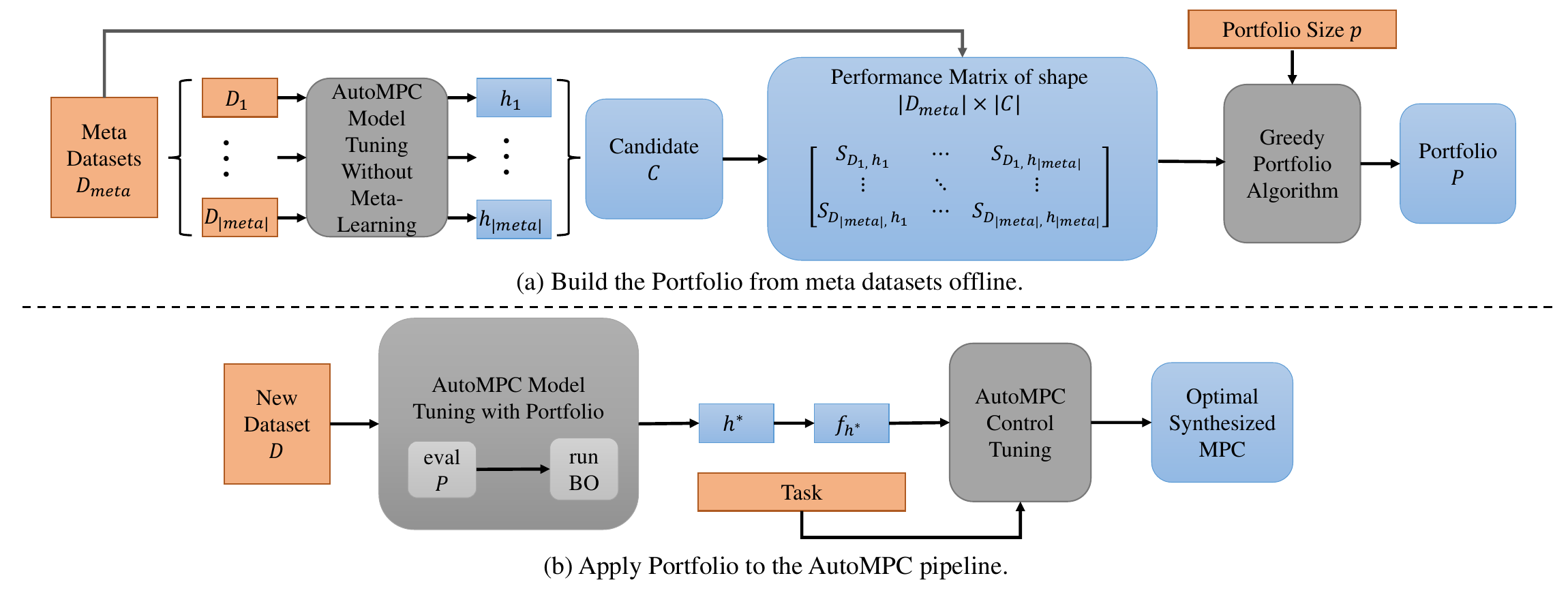}
       \caption{\textbf{Overview of AutoMPC with Portfolio.} Orange rectangular boxes refer to the input data and the blue rounded boxes refer to the output component in each process. (a) We build the Portfolio in a similar way that \cite{feurer-autosklearn} does for AutoML. We first perform AutoMPC model tuning without meta-learning for each meta dataset $D_i$, obtaining the corresponding optimal model configuration $h_i$. These configurations form the candidate set $C$. Then, we evaluate the performance of $C$ on $D_{meta}$ and construct a performance matrix where $S_{D_i,h_j}$ represents the score of configuration $h_j$ on meta dataset $D_i$. Finally, we employ the Greedy Portfolio algorithm proposed by \cite{feurer-autosklearn} to obtain the Portfolio $P$. (b) We utilize $P$ as the initial configurations for BO in AutoMPC model tuning. It will return the best model configuration $h^*$ and correspondent surrogate dynamics model $f_{h^*}$, which will be used for control tuning in AutoMPC and helps to achieve the optimal synthesized MPC.}
      \label{fig:method}
\end{figure*}

%% file: 02-related.tex
\section{Preliminaries}

\subsection{AutoMPC Tuning}

As in \cite{edwards2021automatic}, we consider full-observable, discrete-time dynamical systems of the form 
\begin{equation*}
x_{t+1} = f(x_t, u_t),
\end{equation*}
with state $x_t \in \R^n$ and control $u_t \in \R^m$.  In general, we do not assume access to the ground-truth dynamics $f$, so the function must be approximated with a SysID algorithm to obtain a data-learned dynamics model $\hat{f}$. 


\texttt{AutoMPC} manages design components and hyperparameters using configurations with the \texttt{ConfigSpace} library \cite{config} and conducts Bayesian Optimization (BO) with the \texttt{SMAC3} library \cite{HutHooLey11-smac}. Tuning in \texttt{AutoMPC} can be performed either in a decoupled fashion, where model tuning is performed prior to control tuning, or fully end-to-end, with all hyperparameters being tuned simultaneously. Model tuning\footnote{Note that model tuning and SysID tuning are synonymous in our paper.} only optimizes the SysID hyperparameters for model accuracy, while control tuning optimizes the control performance by tuning the hyperparameters of objective function and optimizer.



\subsection{Meta Learning}


Meta learning, also known as \textit{learning to learn}, is aimed at facilitating effective learning by utilizing prior knowledge of various datasets or tasks \cite{pmlr-v70-finn17a}. It has been applied in the context of MPC to enable fast fine-tuning for unseen tasks \cite{arcari2020}. However, previous work only considers single dynamics model class such as Gaussian process regression, while in our work, we perform automatic selection of the model class. The search space for model tuning is thus very large, including both the choice of model class and the model hyperparameters for each class.
By using meta learning, the efficiency of hyperparameter search can be significantly improved by transferring knowledge between tasks, allowing promising areas in the search space to be quickly identified \cite{feurer2015efficient}. This can accelerate the automatic tuning and synthesis process of the overall \texttt{AutoMPC} pipeline. 



%% file: 03-methods.tex
\section{Method}

Efficiently exploring the extensive search space of potential MPC pipelines is crucial to obtain an effective controller within limited computational resources. \texttt{AutoMPC}~\cite{edwards2021automatic} performs hyperparameter tuning using pure BO, which can be inefficient as it initiates each new problem from scratch and does not take advantage of prior knowledge to identify promising areas in the search space. Pure BO can also cause problem of instability for the tuning process due to the random selection of initial configurations. 

To overcome the limitations of pure BO and enhance the efficiency and effectiveness of \texttt{AutoMPC}, we employ a meta-learning approach called \textit{Portfolio}. This method is inspired by the meta-learning technique proposed in \texttt{Auto-Sklearn 2.0}~\cite{feurer-autosklearn} for AutoML. Portfolio selects a fixed set of configurations from meta datasets to warmstart BO, giving priority to configurations that demonstrate strong performance across a broad range of scenarios. Subsequently, configurations with more specialized applicability are added to the Portfolio. We leverage Portfolio for system ID tuning, which plays a pivotal role in the \texttt{AutoMPC} pipeline.


In the following, we detail the problem statement of SysID tuning in \texttt{AutoMPC} and describe how Portfolio connects to our problem. Figure\ref{fig:method} shows the overview of Portfolio building and how it applies to the \texttt{AutoMPC} pipeline.

\subsection{System ID Tuning Problem Statement}

\texttt{AutoMPC} provides several system ID algorithms which can be used to learn a dynamics model from data. The SysID tuning generates a data-learned dynamics model $\hat{f}_h: (x_t,u_t) \rightarrow x_{t+1}$ with a model configuration\footnote{A model configuration contains a model class and its corresponding hyperparameters.} $h \subseteq \mathcal{H}_S$. The learned dynamics model predicts the state in the next time step based on the current observation and control input.

A nonlinear control benchmark offers a set of state $x_t$ and control $u_t$. Given the number of trajectories $N$ and the length of a trajectory $L$, we can construct a dataset $D=\{ ((x_{i,j}, u_{i,j}), x_{i,j+1}) \}$ where $i \in \{ 0, \ldots, N \}$ and $j \in \{0,\ldots, L-1 \}$, which can be used to learn a dynamics model and evaluate the performance of system ID tuning. 

Given a model configuration $h$ and a dataset $D$, we can empirically approximate the model score\footnote{The smaller the model score, the better the performance.}:

$$
\hat{\epsilon}(h, D) = \frac{1}{N \cdot L} \sum_{i=1}^{N} \sum_{j=1}^{L} \mathcal{L}(\hat{f}_{h}(x_{i,j}, u_{i,j}), x_{i,j+1}),
$$
where $\mathcal{L}$ is the evaluation metric.


During the search, we perform $K$-fold cross-validation to estimate the model score $\hat{\epsilon}_{CV}$ for each model configuration. The goal of SysID tuning is to find an optimal model configuration $h^*$, which can achieve the best model score in the training set:

$$
h^* \in \underset{h \subseteq \mathcal{H}_S}{\text{argmin}} \ \hat{\epsilon}_{CV}(h, D_{train}).
$$ 

\begin{figure}[h!]
    \centering
    \begin{subfigure}{0.45\columnwidth}
      \includegraphics[trim=15 20 10 0, clip, width=\textwidth]{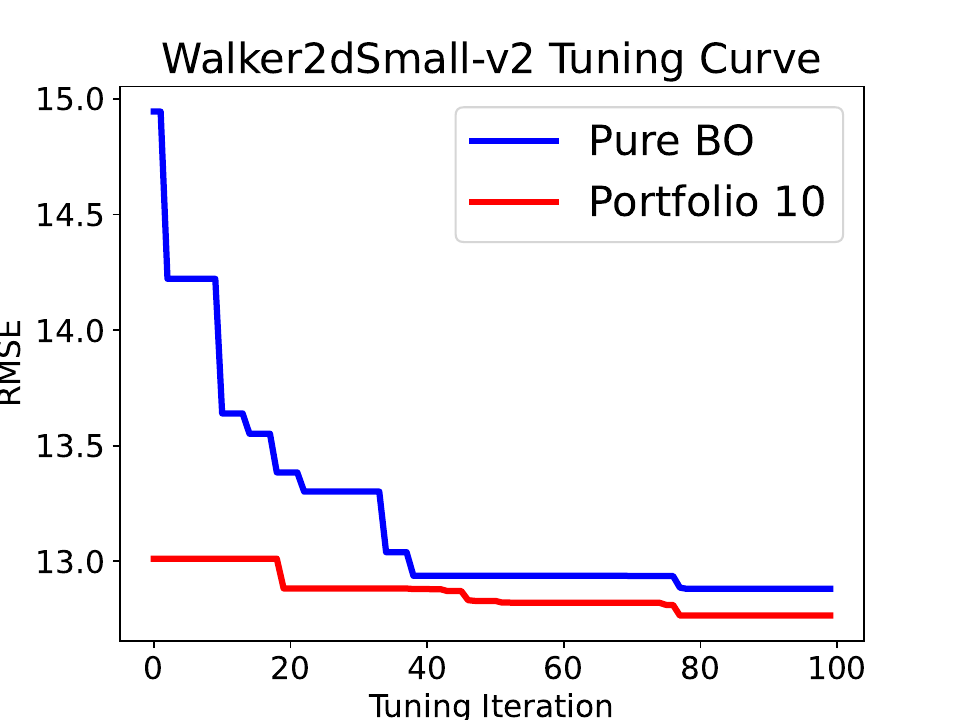}
    \end{subfigure}
    \begin{subfigure}{0.45\columnwidth}
      \includegraphics[trim=15 20 10 0, clip, width=\textwidth]{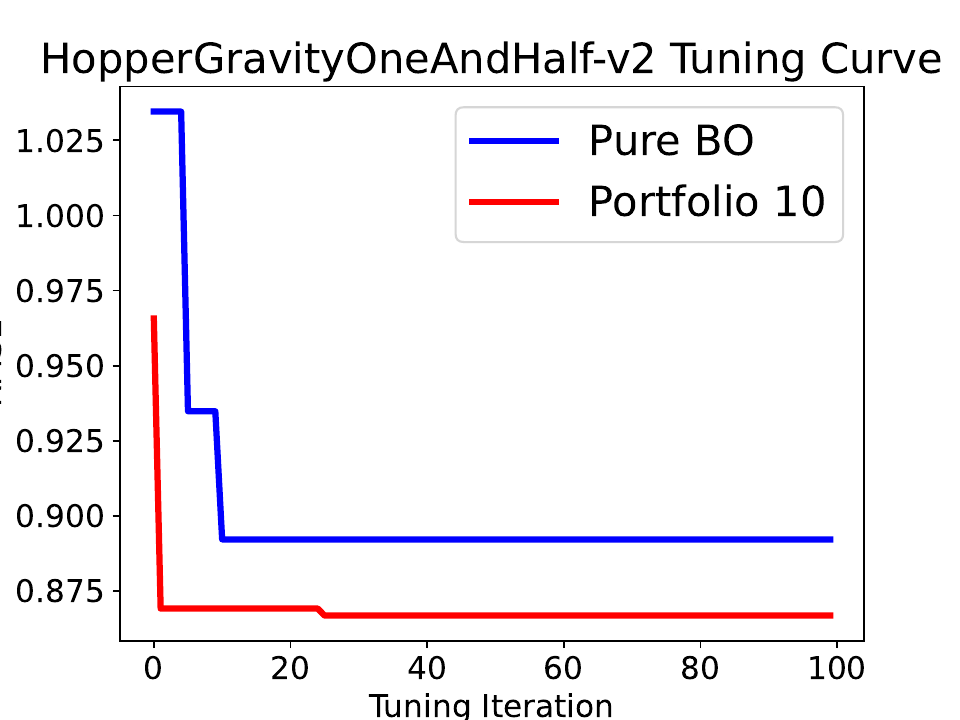}
    \end{subfigure}
     \caption{\textbf{Comparison between Portfolio with size 10 and Pure BO.} We evaluate the model tuning performance on Walker2dSmall and HopperGravityOneAndHalf datasets within 100 iterations. The \textcolor{blue}{blue} lines represent the tuning curve for Pure BO, while the \textcolor{red}{red} lines denote the tuning curve for Portfolio with size 10. Portfolio can lead to a faster convergence of model tuning and outperform pure BO within limited time.}
     \label{fig:port_10}
\end{figure}

\subsection{Portfolio on AutoMPC}

Figure~\ref{fig:method} shows the construction of a Portfolio and how it connects to the \texttt{AutoMPC} system, inspired by \texttt{Auto-Sklearn 2.0}~\cite{feurer-autosklearn} for AutoML. First, we conduct the AutoMPC model tuning separately for each meta dataset without meta-learning. This gives us the optimal configuration for each dataset, forming the candidate set $C$. Next, we evaluate the performance of configurations in $C$ on the meta dataset $D_{meta}$ to create a performance matrix. Using the Greedy Portfolio algorithm proposed by~\cite{feurer-autosklearn}, we construct a Portfolio $P$ by considering the performance matrix and the specified portfolio size. Once we have $P$, we initialize the model tuning using $P$ as the initial configurations for BO. This process yields the best model configuration $h^*$ and its corresponding surrogate dynamics model $f_{h^*}$, which are then used for control tuning in \texttt{AutoMPC}, ultimately helping us achieve the optimal synthesized MPC. Portfolio works by warmstarting BO in SysID tuning and results in a better initialization when encountering new tasks. More details about Greedy Portfolio algorithm can be found in~\cite{feurer-autosklearn}.

%% file: 04-experiments.tex
\section{Experiments and Results}
\label{experiments}


In this section, we investigate the effectiveness of Portfolio on \texttt{AutoMPC}. In particular, we aim to answer the following three questions through the experiments. (1) Can Portfolio enhance the efficiency and stability of SysID tuning? (2) What is the impact of portfolio size $p$ on the tuning performance? (3) Can a superior model obtained through Portfolio lead to a better controller? 



\begin{figure*}
    \centering
    \includegraphics[width=\textwidth]{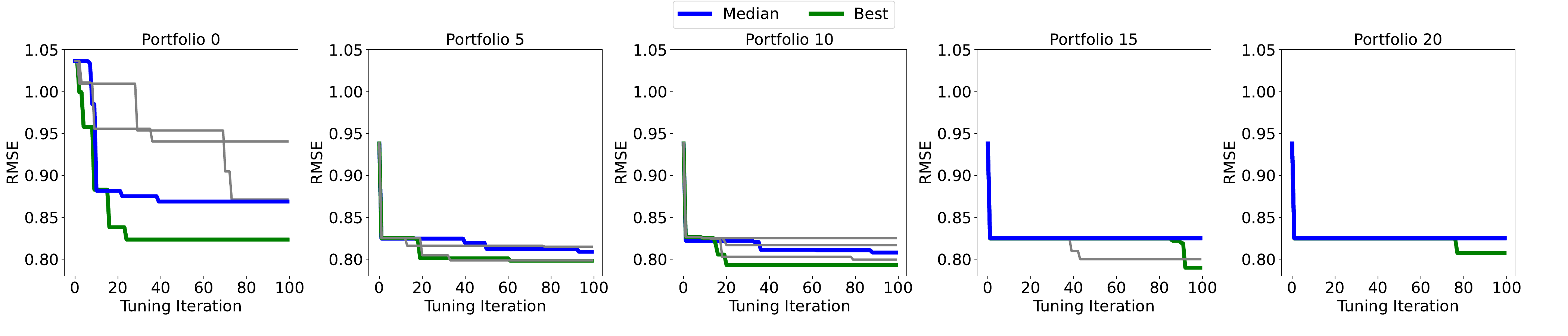}
      \caption{\textbf{Comparison among different Portfolio sizes.} We evaluate the model tuning performance on HopperSmallTorso dataset with pure BO (Portfolio 0) and Portfolio with sizes $5, 10, 15, 20$ on 5 independent runs, plotted in grey. The median and best result are highlighted in blue and green, respectively. In this case, Portfolio consistently outperforms the pure BO regardless of the size. The tuning process becomes more stable as the Portfolio size increases.}
      \label{fig:port_sizes}
\end{figure*}

\begin{figure}
    \centering
      \includegraphics[width=0.8\columnwidth]{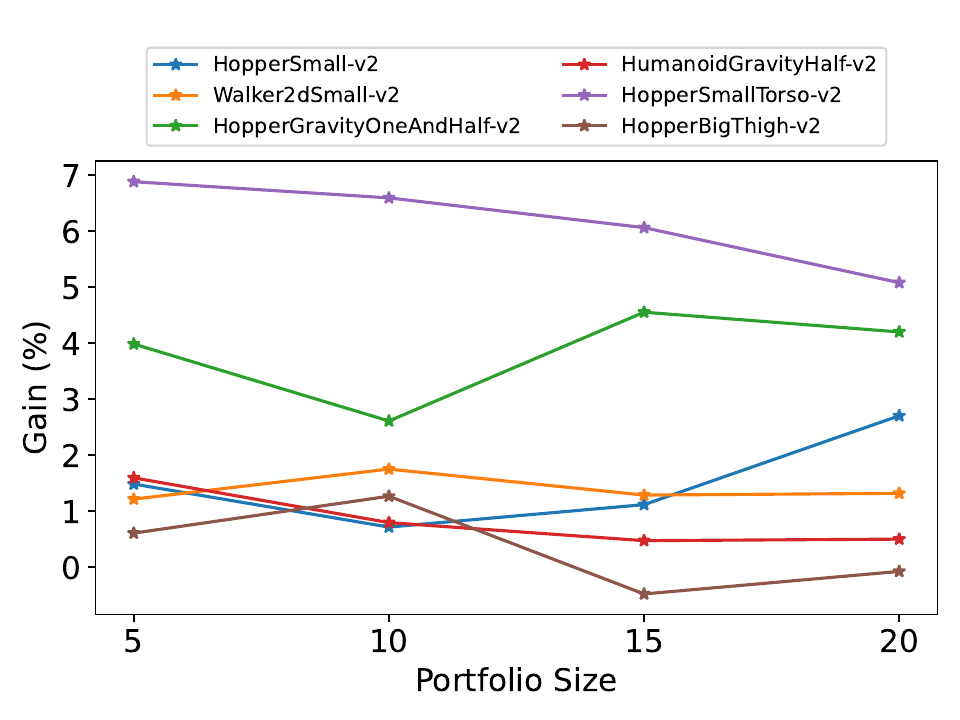}
      \caption{\textbf{Gains among different Portfolio sizes.} We compare the gains of Portfolio with sizes $5, 10, 15, 20$ over the pure BO on 6 datasets. The best Portfolio size varies for different datasets and an inappropriate Portfolio size may worsen the performance.}
      \label{fig:gain}
\end{figure}

\subsection{Benchmarks and Datasets}
\label{exp:benchmark}

We consider benchmark systems and tasks from OpenAI Gym~\cite{gym}, such as Cartpole and HalfCheetah, and their extensions~\cite{henderson2017multitask} by modifying some parameters including gravity and morphological size. The details of our benchmarks can be found in Appendix~\ref{appendix:benchmark}.

We generate each dataset by executing 1,000 trajectories with uniform random controls selected at each time step. Each trajectory has a duration of 10s and a time step of 0.05s. We also include smaller datasets consisting of only 100 trajectories. We choose 20 datasets for Portfolio training and 11 for evaluation. The details of meta training and testing datasets can be found in Appendix~\ref{appendix:datasets}. 


\input{table/port_10}

\subsection{System ID Tuning with Portfolio}
\label{exp:port}

In this study, we aim to evaluate the performance of the learned Portfolio and compare it with the pure BO. 
Each system ID dataset is divided into a training set, a validation set, and a testing set, which contain 70\%, 15\%, and 15\% of the trajectories, respectively. Each method is tuned for 100 iterations using SMAC3 library~\cite{HutHooLey11-smac}, based on the 1 step rollout RMSE prediction error on the validation set.

To account for randomness, we report results averaged over 5 repetitions and also provide the standard deviation over these repetitions. To assess the statistical significance of performance differences, we conduct t-test with $\alpha=0.1$ wherever possible. We set the portfolio size to \textbf{10} and tested our approach on 12 meta testing datasets. We present the results in Table \ref{tb:port_10}. 

As anticipated, Portfolio-based meta-learning generally enhances the model performance within 100 iterations. Additionally, we observe that Portfolio can lead to smaller standard deviations in most cases, improving the stability of the model tuning and reducing the need for multiple attempts to obtain an effective solution.


Furthermore, Figure~\ref{fig:port_10} shows the tuning curves for the median results of 5 independent runs on 2 datasets, which compares the tuning progress between the Portfolio with size 10 and Pure BO. They demonstrate that the initialization provided by Portfolio leads to faster convergence of \texttt{AutoMPC} tuning, which can yield a substantial improvement in model performance when tuning time is limited. For instance, the Walker2dSmall model error can be reduced to below 13 within 20 iterations with Portfolio as opposed to 40 iterations for pure BO. Similarly, for HopperGravityOneAndHalf, Portfolio can achieve a score below 0.875 within 5 iterations, while pure BO cannot achieve this score within 100 iterations. This reduced tuning time can greatly accelerate the \texttt{AutoMPC} workflow since a full 100 iterations of tuning can take more than 24 hours to run.


However, the Portfolio-based meta-learning approach may not be advantageous for out-of-distribution data. The Portfolio is primarily trained on the OpenAI Gym MuJoCo datasets and its extensions, which perform well on datasets with the same data distribution. However, we do not observe any improvement in performance for the out-distribution tasks such as Pendulum and Underwater Soft Robot.

\subsection{Comparison for Different Portfolio Sizes}
\label{exp:size}

In this section, we investigate the impact of portfolio size $p$, a new hyperparameter introduced by Portfolio, on model tuning performance. We conduct experiments with four different portfolio sizes, namely, 5, 10, 15, and 20, and evaluated their performance on the testing datasets.

As depicted in Figure~\ref{fig:port_sizes}, the initial configurations for pure BO approach are random, resulting in large deviation for tuning performance across five attempts. However, Portfolio fixes the initial $p$ configurations and makes the tuning process more stable. The larger the portfolio size is, the more stable the tuning process becomes.

Although the Portfolio with different sizes usually exhibit better performance within a limited number of iterations, there is no correlation between the final result and portfolio size. As shown in Figure~\ref{fig:gain}, the best portfolio size varies depending on the dataset and sometimes an inappropriate portfolio size will worsen the performance of model tuning. 
Therefore, our study shows that portfolio size is an important hyperparameter that affects the stability and effectiveness of the model tuning, which should be selected carefully. 


\subsection{Control Performance with Portfolio}

\input{table/control}

Finally, in this section, we evaluate the impact of model tuning with Portfolio on control performance.  We pick the Cartpole benchmark as a representative task and perform model tuning with portfolio size 0 (Pure BO), 5, and 10, with five seeds each.  We then combine the tuned models with a previously tuned optimizer and objective function to obtain a set of controllers.  (The optimizer and objective hyperparameters were taken from the experiment in Sec.~6.2.1 of \cite{edwards2022thesis}). The average control performance for each portfolio size is reported in Table~\ref{tb:control}.
The control scores range from 0 to 10 with lower being better (see \cite{edwards2021automatic} for details). Both portfolio sizes improve over pure BO and the Portfolio of size 10 achieves the best control score with a 14.76\% improvement over the pure BO. It demonstrates that a superior model obtained through Portfolio can lead to a better controller. 


%% file: table/port_10.tex
\begin{table}[h]
\caption{\textbf{Comparison between Portfolio with size 10 and pure BO on 12 datasets.} Averaged results and standard deviation on 5 independent runs are reported. * indicates the statistically significant differences, with p-value$<0.1$. The best results are in boldface. }
\label{tb:port_10}
\small
\begin{center}
\resizebox{1.0\linewidth}{!}{
\fontsize{20}{30}\selectfont
\begin{tabular}{cccccc}
\hline
\multicolumn{1}{c|}{} & \multicolumn{4}{c|}{\textbf{RMSE}} &  \\
\multicolumn{1}{c|}{} & \multicolumn{2}{c|}{\textbf{BO}} & \multicolumn{2}{c|}{\textbf{Portfolio}} &  \\
\multicolumn{1}{c|}{\multirow{-3}{*}{\textbf{Dataset}}} & \textbf{mean} & \multicolumn{1}{c|}{\textbf{std}} & \textbf{mean} & \multicolumn{1}{c|}{\textbf{std}} & \multirow{-3}{*}{\textbf{Gains (\%)}} \\ \hline
\multicolumn{1}{c|}{HopperSmall} & 0.5987 & \multicolumn{1}{c|}{0.0107} & \textbf{0.5944 } & \multicolumn{1}{c|}{0.0099} & \textit{0.7182} \\
\multicolumn{1}{c|}{Walker2dSmall} & 12.9722 & \multicolumn{1}{c|}{0.2790} & \textbf{12.7447*} & \multicolumn{1}{c|}{0.1073} & \textit{1.7538} \\
\multicolumn{1}{c|}{InvertedPendulumSmall} & 1.2506 & \multicolumn{1}{c|}{0.0555} & \textbf{1.2395 } & \multicolumn{1}{c|}{0.0378} & \textit{0.8876} \\
\multicolumn{1}{c|}{HalfCheetahGravityHalf} & 2.5899 & \multicolumn{1}{c|}{0.0214} & \textbf{2.5499*} & \multicolumn{1}{c|}{\underline{0.0234}} & \textit{1.5445} \\
\multicolumn{1}{c|}{HopperGravityOneAndHalf} & 0.8803 & \multicolumn{1}{c|}{0.0361} & \textbf{0.8573*} & \multicolumn{1}{c|}{0.0171} & \textit{2.6127} \\
\multicolumn{1}{c|}{HumanoidGravityHalf} & 5.5249 & \multicolumn{1}{c|}{0.0849} & \textbf{5.4808 } & \multicolumn{1}{c|}{\underline{0.1070}} & \textit{0.7982} \\
\multicolumn{1}{c|}{HalfCheetahSmallLeg} & 2.5504 & \multicolumn{1}{c|}{0.0340} & \textbf{2.5358 } & \multicolumn{1}{c|}{0.0077} & \textit{0.5725} \\
\multicolumn{1}{c|}{HopperBigThigh} & 0.8194 & \multicolumn{1}{c|}{0.0233} & \textbf{0.8090*} & \multicolumn{1}{c|}{0.0085} & \textit{1.2692} \\
\multicolumn{1}{c|}{HopperSmallTorso} & 0.8656 & \multicolumn{1}{c|}{0.0429} & \textbf{0.8085*} & \multicolumn{1}{c|}{0.0116} & \textit{6.5966} \\ \hline
\multicolumn{6}{c}{\cellcolor[HTML]{329A9D}\textit{\textbf{Out-of-distribution Data}}} \\ \hline
\multicolumn{1}{c|}{Cartpole} & 0.0101 & \multicolumn{1}{c|}{0.0007} & \textbf{0.0097 } & \multicolumn{1}{c|}{0.0012} & \textit{3.9604} \\
\multicolumn{1}{c|}{Pendulum} & \textbf{0.3174} & \multicolumn{1}{c|}{0.0000} & \textbf{0.3174 } & \multicolumn{1}{c|}{0.0001} & \textit{0.0000} \\
\multicolumn{1}{c|}{Underwater Soft Robot} & \textbf{0.1165} & \multicolumn{1}{c|}{0.0000} & \textbf{0.1165 } & \multicolumn{1}{c|}{4.19e-15} & \textit{0.0000} \\ \hline
\end{tabular}}
\end{center}
\end{table}

%% file: table/control.tex
\begin{table}[h]
\caption{Control performance on Cartpole.}
\label{tb:control}
\small
\begin{center}
\resizebox{0.8\linewidth}{!}{
\fontsize{5}{6}\selectfont
\begin{tabular}{c|c|c}
\hline
\textbf{Portfolio Size} & \textbf{Control Score} & \textbf{Gain (\%)} \\ \hline
0 & 8.4000 & - \\
5 & 7.7100 & \textit{8.2143} \\
10 & \textbf{7.1600} & \textit{14.7619} \\ \hline
\end{tabular}}
\end{center}
\end{table}

%% file: 05-conclusion.tex
\section{Conclusion and Future Work}




Our paper is built upon the recent advance of a meta-learning approach named \textit{Portfolio} to enhance the efficiency and robustness of \texttt{AutoMPC}.  We leveraged Portfolio to optimize the initial designs for pure Bayesian Optimization, utilizing a diverse and reliable set of configurations from previous tasks. Our experiments demonstrated that Portfolio significantly improves the capacity of \texttt{AutoMPC} to achieve desirable dynamics models and controllers within limited computational resources.


This work also offers avenues for future extensions. We are interested in exploring Portfolio selection, such as choosing appropriate portfolio size and training datasets, based on the characteristics of particular testing dataset, making \texttt{AutoMPC} more adaptive to out-of-distribution data and open-world robotics.

%% file: 06-appendix.tex
\appendix
\input{table/port_all}
\section{Appendix}

\subsection{Extensive Benchmarks}
\label{appendix:benchmark}

Our extensive benchmark systems are based on OpenAI Gym MuJoCo and their extensions by modifying some parameters. We detail the extensive benchmarks in the following:

\begin{itemize}
    \item \textbf{Open AI Gym MuJoCo:} We consider 10 robotics tasks based on the v-2 environments in Gym MuJoCo including \textit{Ant-v2, HalfCheetah-v2, Hopper-v2, Humanoid-v2, HumanoidStandUp-v2, InvertedDoublePendulum-v2, InvertedPendulum-v2, Reacher-v2, Swimmer-v2, Walker2D-v2, and Pusher-v2. }

    \item \textbf{Modified Gravity:} We adjust the simulated gravity to different scales that range from 0.5 to 1.5 times the normal gravity level, such as \textit{HopperGravityOneAndHalf-v2} and \textit{HalfCheetahGravityHalf-v2}. 
    \item \textbf{Morphological Modifications:} We alter the morphology of a particular body part, such as the foot, leg, thigh, or torso. The body parts that are scaled by 1.25 times their mass and width are labeled as ``Big" while those scaled by 0.75 are classified as ``Small", such as \textit{HopperSmallTorso-v2}. 
    \item \textbf{PusherMovingGoal:} We randomly move the goal position for the Pusher task, namely \textit{PusherMovingGoal-v2}.
\end{itemize}

\subsection{Meta Training and Testing Datasets}
\label{appendix:datasets}

As mentioned in Section~\ref{exp:benchmark}, the datasets are generated by executing trajectories. The normal datasets consists of 1,000 trajectories while the small datasets only have 100 trajectories, such as \textit{HopperSmall-v2} and \textit{Walker2dSmall-v2}.  

We select \textbf{20} datsets for Portfolio training: 

[\textit{HalfCheetah-v2, Hopper-v2, Walker2d-v2, Swimmer-v2, InvertedPendulum-v2, Reacher-v2, Pusher-v2, InvertedDoublePendulum-v2, Ant-v2, Humanoid-v2, HalfCheetahSmall-v2, ReacherSmall-v2, SwimmerSmall-v2, HopperGravityThreeQuarters-v2, Walker2dGravityOneAndHalf-v2, HalfCheetahGravityOneAndQuarter-v2,
HalfCheetahBigThigh-v2, HopperSmallLeg-v2, Walker2dSmallTorso-v2, PusherMovingGoal-v2}].

And \textbf{12} datasets for evaluation:

[\textit{HopperSmall-v2, Walker2dSmall-v2, InvertedPendulumSmall-v2,  
                HalfCheetahGravityHalf-v2, HopperGravityOneAndHalf-v2, HumanoidGravityHalf-v2, 
                HalfCheetahSmallLeg-v2, HopperBigThigh-v2, HopperSmallTorso-v2, 
                CartpoleSwingup, PendulumSwingup, SoftRobot}].

\subsection{More Experimental Results for System ID Tuning with Portfolio}

\begin{figure}[h!]
    \centering
    \begin{subfigure}{0.45\columnwidth}
      \includegraphics[trim=15 20 10 0, clip, width=\textwidth]{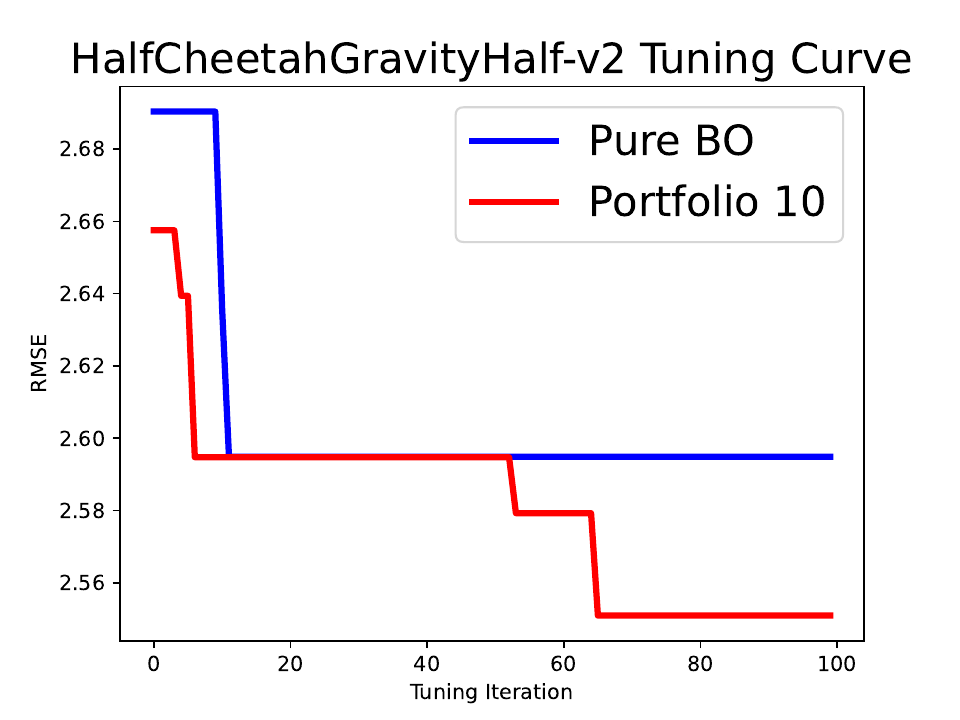}
    \end{subfigure}
    \begin{subfigure}{0.45\columnwidth}
      \includegraphics[trim=15 20 10 0, clip, width=\textwidth]{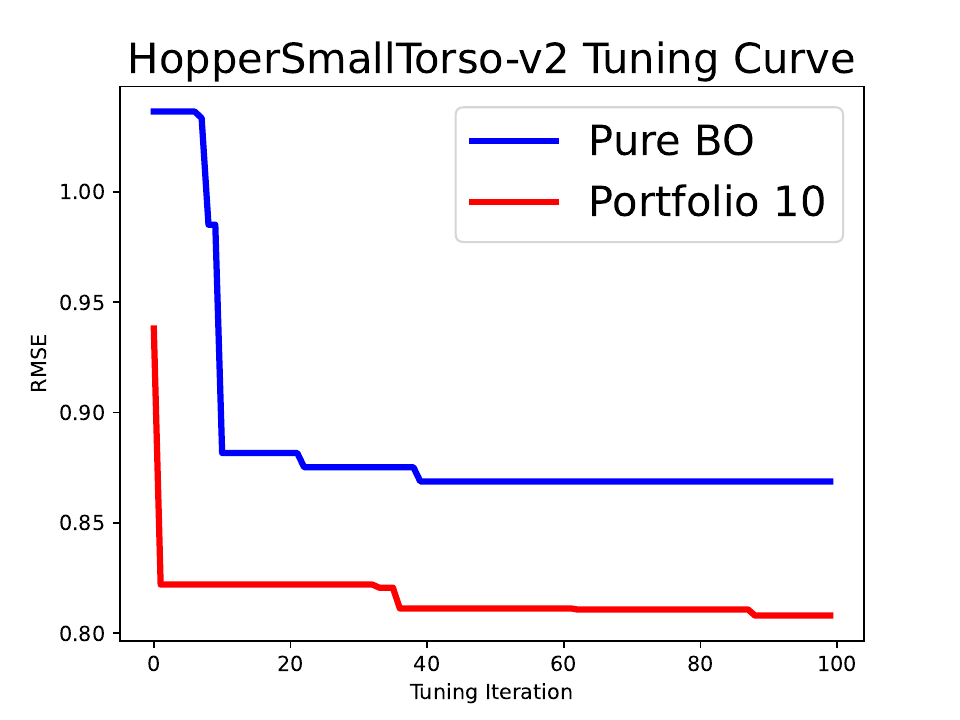}
    \end{subfigure}
    \begin{subfigure}{0.45\columnwidth}
      \includegraphics[trim=15 20 10 0, clip, width=\textwidth]{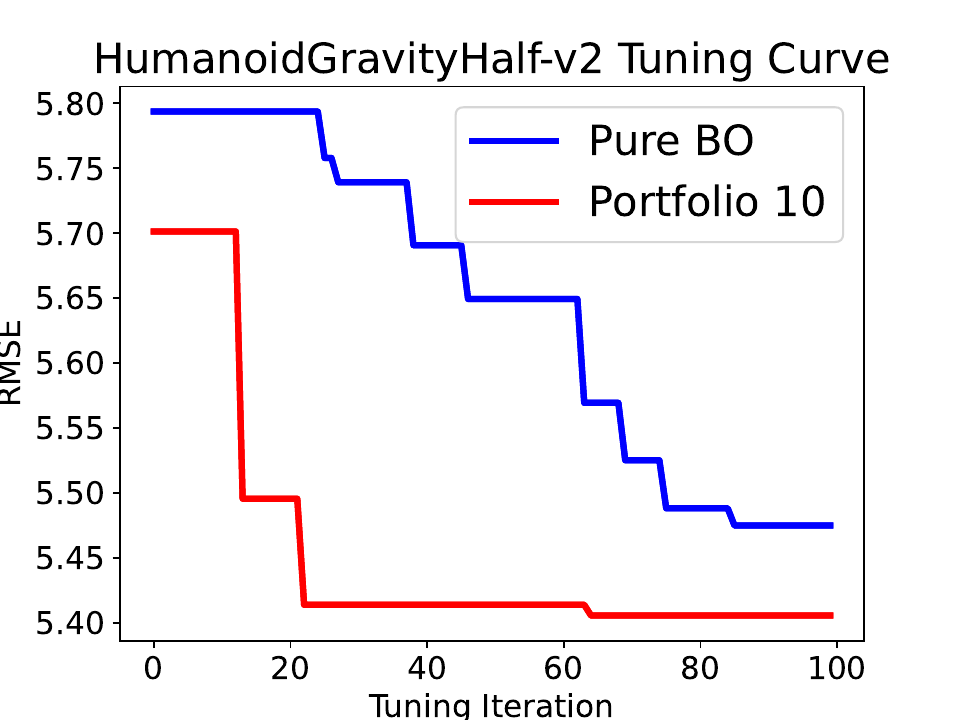}
    \end{subfigure}
    \begin{subfigure}{0.45\columnwidth}
      \includegraphics[trim=15 20 10 0, clip, width=\textwidth]{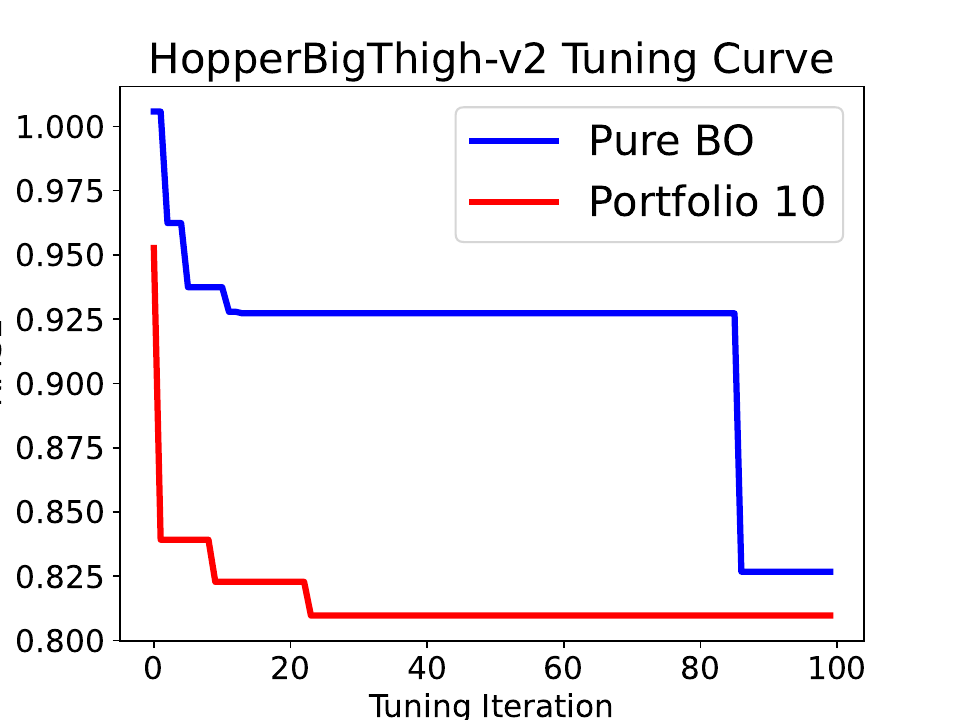}
    \end{subfigure}
     \caption{Tuning curves for Portfolio with size \textbf{10} and Pure BO.}
     \label{fig:port_supp}
\end{figure}

We provide four additional tuning curves for the median results of 5 independent runs to compare Portfolio with 10 and Pure BO in Figure~\ref{fig:port_supp}. They all demonstrate that Portfolio can provide better initialization than the pure BO, which leads to faster convergence of \texttt{AutoMPC} tuning and yields a substantial improvement in model tuning with limited computational resources.

\subsection{More Experimental Results for Different Portfolio Sizes}
\label{appendix:port}

Table~\ref{tb:port_all} shows the results for Portfolio with various sizes (5, 10, 15, 20) and pure BO on 12 testing datasets. For datasets excluding ``Pendulum" and ``Under Soft Robot", Portfolio generally enhances the model performance and stability within 100 iterations. However, the best portfolio size varies depending on the dataset and sometimes an inappropriate portfolio size will worsen the performance of model tuning. 

\subsection{Other Experimental Details}

All experiments are conducted on computer clusters including Illinois Campus Cluster Program\footnote{\url{https://campuscluster.illinois.edu/resources/docs/user-guide/}} and Digital Environments for Learniung, Teaching, and Agency (DELTA)\footnote{\url{https://education.illinois.edu/ci/programs-degrees/delta}}, which consist of 12 compute nodes providing 296 CPU cores and 1152\.GB RAM. 

The experiments also have time-out setting, which will automatically terminate the tuning of each iteration if it exceeds 10 minutes, in which case no result will be provided.

%% file: table/port_all.tex
\begin{table*}[h]
\caption{Comparison between Portfolio with various sizes (5, 10, 15, 20) and pure BO (0) on 12 datasets. Averaged results and standard deviation on 5 independent runs are reported. The best results are in boldface and the second best results are underlined. }
\label{tb:port_all}
\small
\begin{center}
\resizebox{1.0\textwidth}{!}{
\fontsize{20}{35}\selectfont
\begin{tabular}{cccccccccccc}
\hline
\multicolumn{1}{c|}{\textbf{Dataset}} & \multicolumn{10}{c|}{\textbf{RMSE}} &  \\ \cline{1-11}
\multicolumn{1}{c|}{} & \multicolumn{2}{c|}{\textbf{0}} & \multicolumn{2}{c|}{\textbf{5}} & \multicolumn{2}{c|}{\textbf{10}} & \multicolumn{2}{c|}{\textbf{15}} & \multicolumn{2}{c|}{\textbf{20}} &  \\
\multicolumn{1}{c|}{\multirow{-2}{*}{\textbf{Portfolio Size}}} & \textbf{mean} & \multicolumn{1}{c|}{\textbf{std}} & \textbf{mean} & \multicolumn{1}{c|}{\textbf{std}} & \textbf{mean} & \multicolumn{1}{c|}{\textbf{std}} & \textbf{mean} & \multicolumn{1}{c|}{\textbf{std}} & \textbf{mean} & \multicolumn{1}{c|}{\textbf{std}} & \multirow{-3}{*}{\textbf{Best Gains (\%)}} \\ \hline
\multicolumn{1}{c|}{\textbf{HopperSmall}} & 0.5987 & \multicolumn{1}{c|}{0.0107} & {\ul 0.5898} & \multicolumn{1}{c|}{0.0093} & 0.5944 & \multicolumn{1}{c|}{0.0099} & 0.5920 & \multicolumn{1}{c|}{0.0074} & \textbf{0.5825} & \multicolumn{1}{c|}{0.0059} & \textit{2.7059} \\
\multicolumn{1}{c|}{\textbf{Walker2dSmall}} & 12.9722 & \multicolumn{1}{c|}{0.2790} & 12.8142 & \multicolumn{1}{c|}{0.0990} & \textbf{12.7447} & \multicolumn{1}{c|}{0.1073} & 12.8049 & \multicolumn{1}{c|}{0.0317} & {\ul 12.8008} & \multicolumn{1}{c|}{0.0843} & \textit{1.7538} \\
\multicolumn{1}{c|}{\textbf{InvertedPendulumSmall}} & 1.2506 & \multicolumn{1}{c|}{0.0555} & \textbf{1.2182} & \multicolumn{1}{c|}{0.0205} & 1.2395 & \multicolumn{1}{c|}{0.0378} & 1.2775 & \multicolumn{1}{c|}{0.0452} & {\ul 1.2324} & \multicolumn{1}{c|}{0.0470} & \textit{2.5908} \\
\multicolumn{1}{c|}{\textbf{HalfCheetahGravityHalf}} & 2.5899 & \multicolumn{1}{c|}{0.0214} & 2.6418 & \multicolumn{1}{c|}{0.0228} & \textbf{2.5499} & \multicolumn{1}{c|}{0.0234} & {\ul 2.5783} & \multicolumn{1}{c|}{0.0263} & 2.5806 & \multicolumn{1}{c|}{0.0126} & \textit{1.5445} \\
\multicolumn{1}{c|}{\textbf{HopperGravityOneAndHalf}} & 0.8803 & \multicolumn{1}{c|}{0.0361} & 0.8452 & \multicolumn{1}{c|}{0.0033} & 0.8573 & \multicolumn{1}{c|}{0.0171} & \textbf{0.8402} & \multicolumn{1}{c|}{0.0159} & {\ul 0.8433} & \multicolumn{1}{c|}{1.11e-16} & \textit{4.5553} \\
\multicolumn{1}{c|}{\textbf{HumanoidGravityHalf}} & 5.5249 & \multicolumn{1}{c|}{0.0849} & \textbf{5.4366} & \multicolumn{1}{c|}{0.0589} & {\ul 5.4808} & \multicolumn{1}{c|}{0.1070} & 5.4986 & \multicolumn{1}{c|}{0.0549} & 5.4971 & \multicolumn{1}{c|}{0.0400} & \textit{1.5982} \\
\multicolumn{1}{c|}{\textbf{HalfCheetahSmallLeg}} & 2.5504 & \multicolumn{1}{c|}{0.0340} & 2.6279 & \multicolumn{1}{c|}{0.0508} & \textbf{2.5358} & \multicolumn{1}{c|}{0.0077} & {\ul 2.5404} & \multicolumn{1}{c|}{0.0013} & 2.5413 & \multicolumn{1}{c|}{0.0} & \textit{0.5725} \\
\multicolumn{1}{c|}{\textbf{HopperBigThigh}} & 0.8194 & \multicolumn{1}{c|}{0.0233} & {\ul 0.8144} & \multicolumn{1}{c|}{0.0084} & \textbf{0.8090} & \multicolumn{1}{c|}{0.0085} & 0.8233 & \multicolumn{1}{c|}{0.0002} & 0.8200 & \multicolumn{1}{c|}{0.0041} & \textit{1.2692} \\
\multicolumn{1}{c|}{\textbf{HopperSmallTorso}} & 0.8656 & \multicolumn{1}{c|}{0.0429} & \textbf{0.8060} & \multicolumn{1}{c|}{0.0066} & {\ul 0.8085} & \multicolumn{1}{c|}{0.0116} & 0.8131 & \multicolumn{1}{c|}{0.0151} & 0.8216 & \multicolumn{1}{c|}{0.0071} & \textit{6.8854} \\ \hline
\multicolumn{12}{c}{\cellcolor[HTML]{329A9D}\textit{\textbf{Out-of-distribution Data}}} \\ \hline
\multicolumn{1}{c|}{\textbf{Cartpole}} & 0.0101 & \multicolumn{1}{c|}{0.0007} & \textbf{0.0088} & \multicolumn{1}{c|}{0.0012} & {\ul 0.0097} & \multicolumn{1}{c|}{0.0012} & 0.0101 & \multicolumn{1}{c|}{0.0003} & {\ul 0.0097} & \multicolumn{1}{c|}{0.0011} & \textit{12.8713} \\
\multicolumn{1}{c|}{\textbf{Pendulum}} & \textbf{0.3174} & \multicolumn{1}{c|}{0.0} & \textbf{0.3174} & \multicolumn{1}{c|}{4.15e-16} & \textbf{0.3174} & \multicolumn{1}{c|}{0.0001} & \textbf{0.3174} & \multicolumn{1}{c|}{6.25e-05} & \textbf{0.3174} & \multicolumn{1}{c|}{5.75e-06} & \textit{0} \\
\multicolumn{1}{c|}{\textbf{Underwater Soft Robot}} & \textbf{0.1165} & \multicolumn{1}{c|}{4.19e-15} & \textbf{0.1165} & \multicolumn{1}{c|}{0.0001} & \textbf{0.1165} & \multicolumn{1}{c|}{4.19e-15} & \textbf{0.1165} & \multicolumn{1}{c|}{0.0001} & \textbf{0.1165} & \multicolumn{1}{c|}{4.19e-15} & \textit{0} \\ \hline
\end{tabular}}
\end{center}
\end{table*}